\documentclass[lettersize,journal]{IEEEtran}
\usepackage{listings}
\usepackage{amssymb}
\usepackage[dvipsnames]{xcolor} 
\usepackage{hyperref}
\usepackage{graphicx}
\usepackage{multirow}
\usepackage{xurl}
\usepackage{lipsum}
\usepackage{lscape}
\usepackage{amsmath}
\usepackage{relsize}
\usepackage[figuresright]{rotating}
\usepackage{lineno}
\usepackage{algorithm}
\usepackage{algpseudocode}
\usepackage{comment}
\usepackage{nomencl}
\makenomenclature
\usepackage{calc}
\usepackage[T1]{fontenc}
\usepackage[caption=false,font=footnotesize]{subfig}
\usepackage{tikz}
\usetikzlibrary{fit}
\usepackage{float}
\usepackage[section]{placeins}
\usepackage{orcidlink}
\usepackage{setspace}
\usepackage{tabularx}
\usepackage{array}
\usepackage{makecell}
\usepackage{threeparttable}
\usepackage{booktabs}
\hypersetup{
  colorlinks,
  citecolor=Violet,
  linkcolor=Red,
  urlcolor=Blue}

\begin{document}
\title{Agibot: An Open-Source Modular Swerve-Drive Robot Platform for Agricultural Robotics Research and Education}
\author{Moeen Ul Islam$^1$, Bishal Adhikari$^1$, Kori Edwards$^1$, Lorien Harvey$^1$, J. Alex Thomasson$^{1, 2}$ and Dong Chen$^{1*}$%
\thanks{$^1$Moeen Ul Islam, Bishal Adhikari, Kori Edwards, Lorien Harvey, and Dong Chen are with the Department of Agricultural and Biological Engineering, Mississippi State University, Mississippi State, MS 39762, USA. Emails: {\tt\{mu136, ba974, kbe77, llh320, dc2528\}@msstate.edu}.}%
\thanks{$^{1,2}$J. Alex Thomasson is with the Agricultural Autonomy Institute and the Department of Agricultural and Biological Engineering, Mississippi State University, Mississippi State, MS 39762, USA. Email: \texttt{athomasson@aai.msstate.edu}.}
\thanks{$^*$Dong Chen is the corresponding author.}
\thanks{This work has been submitted to the IEEE for possible publication. Copyright may be transferred without notice, after which this version may no longer be accessible.}}

\maketitle

\begin{abstract}
Agricultural production faces persistent labor shortages and increasing demand for automation, yet progress in agricultural robotics remains constrained by the limited availability of affordable, flexible, and accessible research platforms. This paper presents Agibot, an open-source, modular, low-cost agricultural robotics platform designed for autonomy research and experimentation, with potential applications in hands-on robotics education. The platform integrates four-wheel independent swerve drive, swappable computing, distributed embedded control, multimodal sensing, and configurable manipulation within a unified sense-plan-act-verify framework supporting perception, control, navigation, manipulation, and autonomous task verification.
The platform was evaluated through representative row-crop navigation and manipulation tasks under varying lighting and soil conditions. Experiments demonstrated reliable lane following, autonomous row-to-row navigation, target identification, seed dispensing, plant manipulation, and closed-loop verification of task success. Its modular architecture supports independent testing, modification, and validation of autonomy components for rapid prototyping and experimentation. These proof-of-concept tasks, controller-tuning studies, and integrated trials demonstrate the platform's research capabilities, while its accessible and reconfigurable design supports its use for hands-on robotics education.
By combining affordability, modularity, open-source accessibility, configurable manipulation, and end-to-end sense-plan-act-verify autonomy, Agibot provides a practical platform for agricultural robotics research, technology development, and workforce training.
%The design won first place at ASABE xx.
\end{abstract}


\section{Introduction}
\label{sec:introduction}
Agricultural production continues to face persistent labor shortages, rising labor costs, and increasing pressure to improve productivity and sustainability. These challenges have accelerated interest in autonomous robotic systems capable of performing agricultural tasks such as navigation, monitoring, precision intervention, and manipulation in complex and highly variable field environments \cite{martin2022robots,bazargani2024automation}. Unlike structured industrial settings, agricultural environments contain changing lighting conditions, irregular terrain, biological variability, and dynamic obstacles, requiring robots to do more than execute pre-programmed motions \cite{bechar2016agricultural,bechar2017agricultural}. Effective agricultural robots must continuously sense their surroundings, make decisions, perform actions, and verify that those actions were successfully completed.

Despite rapid advances in agricultural robotics, progress remains constrained by the limited availability of affordable, flexible, and accessible research platforms \cite{kemeshi2026,nasika2026}. Commercial agricultural robotic platforms with integrated autonomy are often proprietary and relatively costly, with limited flexibility for hardware and software modification, which can constrain their use in research and education \cite{xu2022mars,jiang2026plantprotection}.
This challenge affects not only researchers seeking to develop and validate new autonomy algorithms but also educators training the next generation of agricultural robotics engineers. 
While several low-cost robotic platforms have been developed for research and education, most remain focused on individual platforms, subsystems, or task-specific capabilities, with comparatively limited emphasis on unified frameworks integrating perception, decision-making, control, and task execution \cite{gu2026heterogeneous}. Furthermore, most existing educational platforms are designed for indoor environments and lack validation in agricultural settings characterized by variable lighting and unstructured terrain \cite{nwankwo2023romr}.

\begin{figure*}[t]
  \centering
  \includegraphics[width=0.85\textwidth]{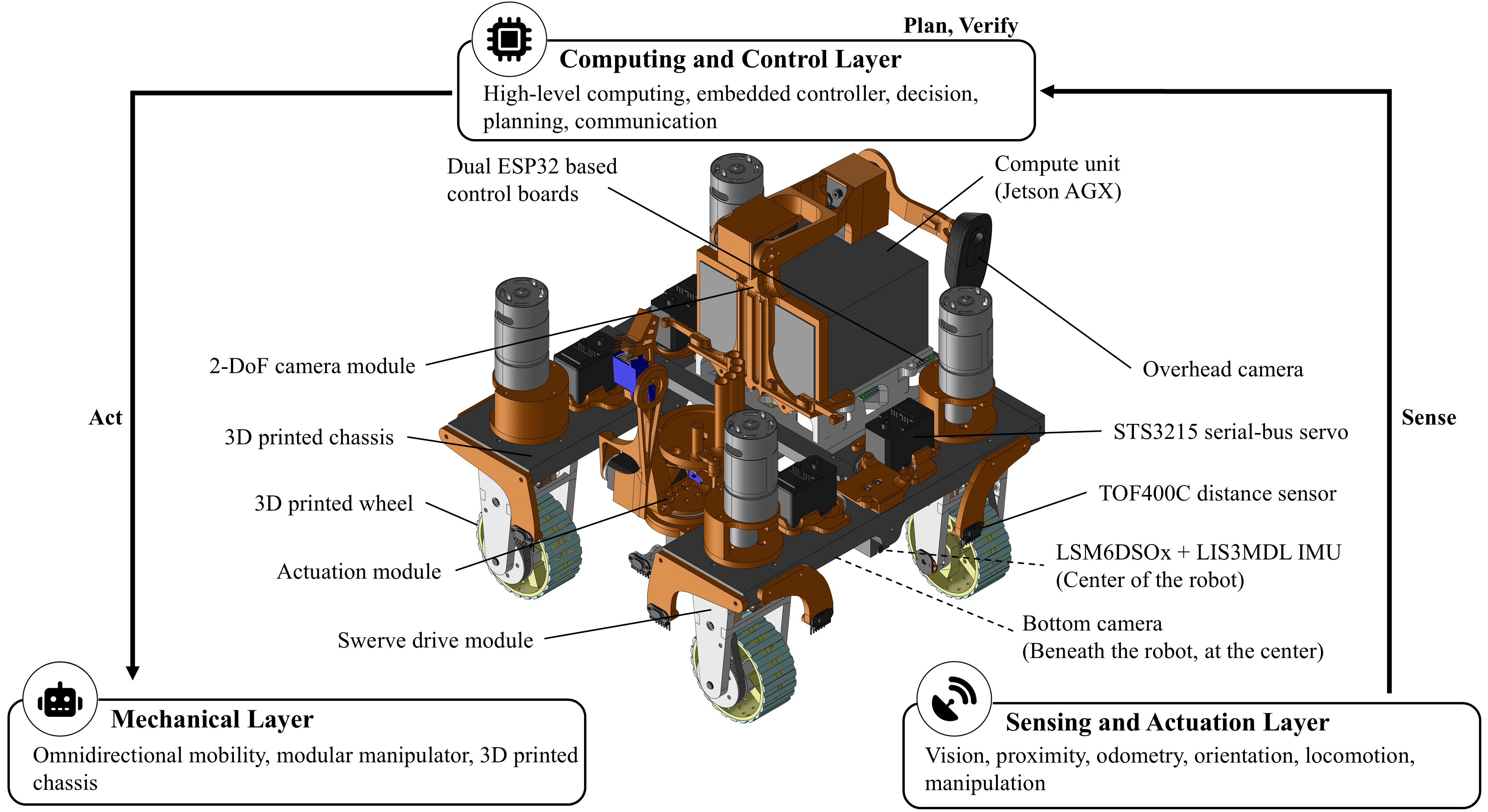}
  \caption{Overview of the Agibot platform showing major components mapped to the Computing and Control, Sensing and Actuation, and Mechanical layers, connected via the \textit{Sense-Plan-Act-Verify} control loop.}
  \label{fig:chassis}
\end{figure*}

To address these challenges, this paper presents Agibot (Fig.~\ref{fig:chassis}), an open-source and modular robotic platform for agricultural robotics research and workforce development. The platform combines a four-wheel independent swerve-drive chassis, distributed embedded control, swappable computing hardware, multimodal sensing, and a configurable manipulation system. Its autonomy architecture follows a unified \textit{sense-plan-act-verify} framework, in which the robot senses its environment, plans an action, executes the task, and verifies the resulting outcome.

The modular architecture supports both subsystem-level experimentation and integrated autonomy development. Researchers can independently modify and evaluate perception, control, navigation, manipulation, and computing components without redesigning the complete platform, while learners can develop and integrate these same autonomy elements through hands-on experimentation. This combination of reconfigurable hardware, accessible software interfaces, and end-to-end autonomy support makes the platform suitable for both research and classroom demonstrations. This paper makes the following primary contributions:
\begin{enumerate}
   \item An open-source, modular agricultural robotics platform featuring independent swerve mobility, distributed embedded control, multimodal sensing, swappable computing hardware, and configurable manipulation capabilities for agricultural autonomy research and education. Hardware designs and software are publicly available at \url{https://moeen14.github.io/agibot-homepage/}.
   \item Two representative proof-of-concept manipulation tasks, i.e., autonomous seed dispensing and plant toppling, are implemented as
representative agricultural applications to demonstrate the platform's
manipulation capabilities.
   \item A unified \textit{sense-plan-act-verify} autonomy framework that supports both subsystem-level development and end-to-end autonomous operation, including perception, navigation, manipulation, and task verification.
   \item Experimental validation of the integrated platform in a representative agricultural testbed under variable lighting and soil/terrain conditions.
\end{enumerate}

The remainder of this paper is organized as follows. Section~\ref{sec:related-work} reviews related open-source robotic platforms. Section~\ref{sec:platform-design} describes the mechanical, sensing, actuation, and computing architecture of Agibot. Section~\ref{sec:autonomy-methods} presents the autonomy methodology, including navigation, manipulation, and closed-loop verification. Section~\ref{sec:experimental-evaluation} reports the experimental setup and system evaluation results. Finally, Section~\ref{sec:conclusion} summarizes the findings and outlines future research directions. Additional experimental analyses are provided in the Appendix.

\section{Open-Source Robotics Platforms for Research and Education}
\label{sec:related-work}

\begin{table*}[t] 
\centering 
\begin{threeparttable} 
\caption{Comparison of representative open-source mobile robotic platforms. The comparison distinguishes available hardware from autonomy functions demonstrated as part of an integrated robotic workflow.} 
\label{tab:platform-comparison} 
\footnotesize \setlength{\tabcolsep}{3.5pt} \renewcommand{\arraystretch}{1.15} \begin{tabularx}{\textwidth}{ >{\raggedright\arraybackslash}p{1.75cm} >{\raggedright\arraybackslash}p{2.25cm} >{\raggedright\arraybackslash}p{2.55cm} >{\raggedright\arraybackslash}X >{\raggedright\arraybackslash}p{2.15cm} >{\raggedright\arraybackslash}p{2.35cm} >{\centering\arraybackslash}p{1.35cm} } 
\toprule 
\textbf{Platform} & \textbf{Mobility} & \textbf{Onboard Computing} & \textbf{Primary Sensors} & \textbf{Manipulation} & \textbf{Validation Environment} & \textbf{Demonstrated Functions} \\ 
\midrule 
Thymio \cite{mondada2017thymio} & Differential drive & PIC24F MCU & IR proximity and ground sensors & None & Desktop and tabletop & P, C \\ 
\midrule 
Mona \cite{arvin2019mona} & Differential drive & ATmega328; optional Pi Zero & Five IR proximity sensors & None & Indoor arena & P, C \\ \midrule 
SMARTmBOT \cite{jo2022smartmbot} & Differential drive with chain-linked front and rear wheels & Raspberry Pi 4; ROS~2 & Eight ToF sensors, camera, and IR line sensors & Optional 5-DOF arm & Indoor laboratory with motion-capture tracking & P, C \\ 
\midrule 
ROBOTONT \cite{raudmae2023robotont} & Three-wheel Kiwi omnidirectional drive & Intel NUC and STM32 & RealSense RGB-D camera & None & Indoor classroom and laboratory & P, C, Pl \\ 
\midrule 
ROMR \cite{nwankwo2023romr} & Differential drive with hoverboard BLDC motors & Jetson Nano and Arduino Mega & RGB-D camera, tracking camera, LiDAR, and IMU & None & Indoor environment & P, C, Pl\tnote{a} \\ 
\midrule MBot \cite{gaskell2024mbot} & Differential, omnidirectional, or Ackermann drive & Raspberry Pi 4 or Jetson Nano; tier-dependent & LiDAR and camera; tier-dependent & None & Indoor courses and laboratories & P, C, Pl \\ 
\midrule 
MecQaBot \cite{james2024mecqabot} & Differential or omnidirectional drive & Raspberry Pi 4B; optional Arduino & Camera, LiDAR, and 9-DOF IMU & None & Indoor teaching laboratory & P, C \\ 
\midrule 
PlatROB \cite{balbuena2026platrob} & Ackermann, omnidirectional, or differential drive & Jetson Nano and module-level Arduino & Ultrasonic sensors and IMU; optional camera and LiDAR & Optional 4-DOF arm & Structured indoor track & P, C, Pl \\ 
\midrule 
Agibot & Four-wheel independent swerve drive (8 DOF); skid-steer mode & Swappable; Jetson AGX Orin and dual ESP32 controllers & Dual actuated cameras, ToF sensors, IMU, and wheel encoders & Swappable 3-DOF hybrid manipulator & Unstructured, soil-covered agricultural testbed & P, C, Pl, M, V \\ 
\bottomrule 
\end{tabularx} 
\begin{tablenotes}[flushleft] 
\footnotesize 
\item P = perception; C = control; Pl = planning; M = manipulation; V = post-action task verification. 
\item[a] Partial planning functionality was demonstrated. \end{tablenotes} 
\end{threeparttable} 
\end{table*}

The growing demand for autonomous agricultural systems has created a corresponding need for accessible robotic platforms that support both technology development and practical training. Open-source and low-cost robotics platforms have become important enablers of robotics research and education because they reduce development costs, improve accessibility, and allow researchers and students to modify both hardware and software to meet specific objectives \cite{pereira2026rosenabled}. Recent reviews have identified a rapidly expanding ecosystem of ROS-enabled mobile robots developed for higher education, research laboratories, and robotics competitions, reflecting increasing efforts to democratize robotics research and training \cite{pereira2026rosenabled}. Table~\ref{tab:platform-comparison} summarizes representative open-source mobile robotic platforms and compares their hardware, validation environments, and demonstrated autonomy functions.

Existing open-source platforms vary substantially in their hardware configurations, autonomy capabilities, and intended users. Educational robots such as Thymio and Mona focus primarily on introductory robotics concepts using microcontroller-based architectures and basic sensor suites \cite{mondada2017thymio, arvin2019mona}. More advanced platforms, including SMARTmBOT, MecQaBot, ROBOTONT, and MBot, integrate onboard computing, cameras, LiDAR, and ROS-based software frameworks to support university-level instruction in perception, navigation, mapping, and autonomous control \cite{jo2022smartmbot, james2024mecqabot, raudmae2023robotont,gaskell2024mbot}. Although these systems provide valuable tools for robotics education and algorithm development, their evaluation has generally focused on structured indoor environments.

Several open-source platforms have extended these capabilities toward research-oriented applications. ROMR integrates Jetson-based computing, RGB-D sensing, visual tracking, and LiDAR for autonomous navigation research \cite{nwankwo2023romr}. PlatROB uses a modular architecture that supports multiple drivetrain configurations, hardware arrangements, and an optional educational manipulator \cite{balbuena2026platrob}. In agricultural robotics, platforms such as BoniRob and Thorvald II have demonstrated the value of reconfigurable robotic systems for field operations, crop monitoring, and individual-plant treatment \cite{ruckelshausen2009bonirob, grimstad2017thorvald}. However, these agricultural platforms were developed primarily as full-scale research machines rather than low-cost, readily reproducible systems for combined research and instruction.

Despite these advances, existing open-source platforms generally address selected autonomy functions or are evaluated in structured settings without the lighting, surface, and task variability representative of agricultural robotics. Integrated systems combining modular mobility, perception, navigation, manipulation, and closed-loop task verification in an affordable, reproducible platform remain comparatively limited \cite{Liu2024}. Although some platforms provide optional manipulators, they generally do not demonstrate an end-to-end workflow that perceives a target, plans and executes a physical action, verifies the outcome, and responds to unsuccessful execution. Agibot addresses this gap by integrating reconfigurable hardware, swappable computing, autonomous navigation, configurable manipulation, and post-action verification within one open-source platform. Its complete workflow is evaluated under variable lighting and soil conditions representative of agricultural operation.

\section{Platform Design}
\label{sec:platform-design}

\begin{figure*}[!t]
  \centering
  \includegraphics[width=0.95\textwidth]{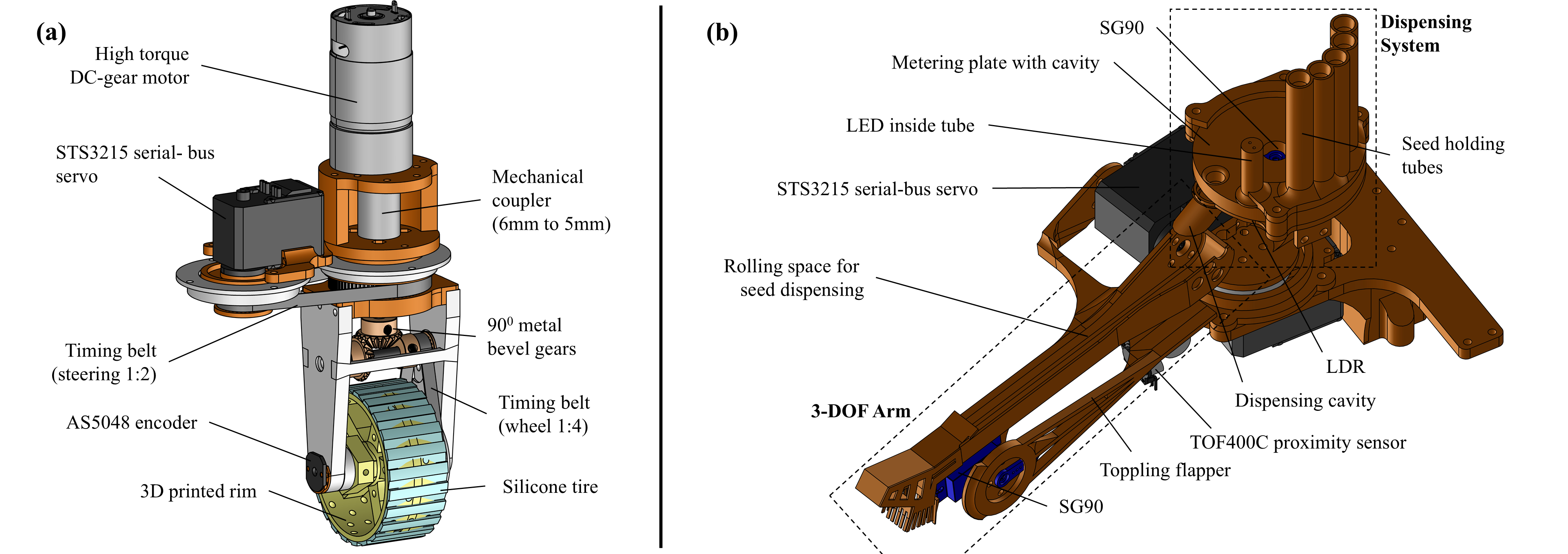}
  \caption{(a) A single swerve module: an
  STS3215 steering servo drives azimuth through a 1:2 timing belt, while a high-torque DC gear motor drives the
  wheel through a mechanical coupler, a $90^\circ$ bevel-gear pair, and a 1:4 timing belt, with an AS5048 encoder
  at the wheel hub. (b) The hybrid 3-DOF manipulation arm and its seed-dispensing system.}
  \label{fig:mechanism-detail}
\end{figure*}

Agibot is a compact, self-contained mobile robot with onboard computing, sensing, and manipulation hardware, requiring no external infrastructure for operation. Figure~\ref{fig:chassis} illustrates the overall system architecture, which is organized into three interconnected layers: (1) a mechanical layer providing mobility, structural support, and standardized mounting interfaces; (2) a sensing and actuation layer supporting environmental perception and physical interaction; and (3) a computing and control layer implementing distributed hardware control and autonomous decision-making. This layered architecture allows individual subsystems to be modified or replaced while maintaining compatibility with the complete autonomy framework. 

\subsection{Mechanical design}
\label{sec:mechanical}
Agibot has an overall footprint of $11.84 \times 11.88~\mathrm{in.}^{2}$, a height of $11.86~\mathrm{in.}$, and a total mass of $4.58~\mathrm{kg}$, including the Jetson AGX Orin. The platform, excluding the swappable high-level computer, costs approximately \$680; the evaluated configuration with a \$1,000 Jetson AGX Orin 64~GB therefore costs approximately \$1,680. Based on computing-unit prices at the time of development, estimated system totals are \$1,080 with a \$400 Jetson Orin Nano, \$870 with an 8-GB Raspberry Pi 5 and 26-TOPS AI HAT+ (\$190 combined), and \$760 with the Raspberry Pi 5 alone. The lower-cost devices were evaluated for individual perception models but not for concurrent execution of the complete autonomy stack. These first-build costs prioritize off-the-shelf availability and reproducibility; design and component optimization could further reduce cost.

The chassis is composed primarily of 3D-printed polylactic acid (PLA) components reinforced by three 5-mm-diameter steel rods. Threaded inserts and distributed mounting features are incorporated throughout the printed structure to simplify the attachment, removal, and repositioning of sensor brackets, electronic modules, camera assemblies, and other hardware. The batteries, power-distribution hardware, motor drivers, and embedded controllers are housed within the central compartment, while the upper frame supports the high-level computer and actuated camera assembly. This arrangement maintains a low center of mass, minimizes cable routing, and provides convenient access to the electronic components.

To support precise lateral positioning, heading correction, and maneuvering in confined agricultural environments, the platform uses four independently controlled swerve-drive modules positioned around the chassis, shown in Fig.~\ref{fig:mechanism-detail}(a). Together, the modules provide eight actively controlled degrees of freedom (DOF) through four wheel-drive axes and four steering axes. Independent regulation of wheel velocity and steering azimuth enables simultaneous lateral and heading correction, fine positioning near manipulation targets, and operation in confined spaces. The drivetrain can also operate in skid-steer mode by fixing the steering azimuths and independently controlling the left- and right-side wheel velocities. Switching between the two mobility modes is achieved through software commands without mechanical reconfiguration.

Each swerve module, contains an STS3215 serial-bus servo for steering and a 12-V DC gearmotor for wheel propulsion. The steering servo drives the wheel assembly through a 2:1 timing-belt reduction, while the drive motor transfers power through a coaxial shaft, a 90-degree bevel-gear pair, and a 4:1 timing-belt reduction. An AS5048A magnetic encoder mounted at each wheel hub provides direct rotational feedback for wheel-speed estimation and motion monitoring. Each wheel consists of a 3D-printed rim and a 2-mm-thick soft silicone tire formed by curing silicone around the rim. %The swerve-module and drivetrain configuration is shown in Fig.~\ref{fig:mechanism-detail}(a).

To support multiple physical interaction tasks without requiring separate task-specific manipulators, the platform incorporates a hybrid manipulation module, shown in Fig.~\ref{fig:mechanism-detail}(b), that combines a 3-DOF arm with an integrated dispensing mechanism. The dispensing unit uses vertically mounted storage tubes and a rotating metering plate with a single cavity that transfers one item at a time from the storage positions, through an intermediate verification position monitored by an LED-LDR pair mounted above and below the cavity, to a dispensing cavity. From there, the item is guided through an internal channel along the arm and released at the distal end. The 3-DOF arm provides adjustable positioning of the shared end-effector region and supports multiple interaction modes through the same mechanical structure, including dispensing and contact-based manipulation.

%The hybrid module and dispensing configuration are shown in Fig.~\ref{fig:mechanism-detail}(b).

The sensing hardware is distributed around the chassis according to its mechanical function. The lower camera is mounted beneath the chassis on a rotatable and manually adjustable interface actuated by an STS3215 servo. The upper camera is mounted on a 2-DOF camera module, an actuated mechanism that provides adjustable height and viewing orientation. The module allows the camera to remain compact in its lowered configuration and to extend upward to increase scene coverage. Seven TOF400C modules are positioned around the platform for perimeter and target ranging, with one mounted directly beneath the lower camera to provide distance information along the camera viewing direction. An LSM6DSOx accelerometer-gyroscope and an LIS3MDL magnetometer form the centrally mounted IMU, while AS5048A magnetic encoders are integrated at the wheel hubs and the STS3215 servos incorporate encoders within the steering and manipulator joints.

The open-source, 3D-printable mechanical design allows the chassis and standardized mounting features to be readily modified for alternative wheel modules, sensors, cameras, computing hardware, and manipulators. The manipulator is attached through a removable interface that can accommodate alternative arms and end effectors. This modular construction enables the platform to be reconfigured for different agricultural robotics research and instructional activities while preserving a common mechanical interface for the sensing and actuation hardware described next.

\subsection{Sensing and actuation}
\label{sec:sensors}

The platform integrates visual, ranging, inertial, and proprioceptive sensing for navigation, manipulation, and execution monitoring. Two Logitech C270 cameras provide complementary views of the operating environment. Once the robot reaches a candidate target, it performs \textit{task detection}: a deep-learning model identifies the target configuration in front of the robot and determines the required response %-- plant toppling, seed dispensing, or no action, in which case the robot continues navigating to the next instance -- 
and pre-actuation alignment, allowing the system to determine an appropriate response and position the manipulator relative to the target. The upper camera mounted on the camera module supports navigation and post-actuation \textit{task validation}, where visual feedback is used to assess the target state and determine whether the intended action was successfully completed.

Seven TOF400C modules based on the VL53L1X time-of-flight (ToF) sensor provide range information for obstacle detection, boundary monitoring, target approach, and final positioning. The LSM6DSOx accelerometer-gyroscope and LIS3MDL magnetometer provide platform-state information, including orientation, angular velocity, linear acceleration, and magnetic heading. AS5048A magnetic encoders provide wheel-rotation feedback for odometry and motion verification, allowing the system to detect cases in which commanded wheel motion is not achieved. The internal encoders of the STS3215 servos provide steering-angle feedback for omnidirectional locomotion, manipulator joint-state feedback, and position feedback for the actuated camera module.

%To demonstrate the platform's capability to perform physical tasks relevant to agricultural robotics, two representative proof-of-concept operations were implemented: toppling selected plant surrogates and dispensing individual seed surrogates. These simplified tasks were not intended to reproduce complete field operations. Instead, they provide repeatable examples for evaluating target detection, platform alignment, manipulation, and task-outcome verification within an integrated autonomy workflow.

Physical interaction is performed through the 3-DOF manipulator and integrated dispensing mechanism. Two STS3215 servos control arm-base rotation and arm bending, while an SG90 micro servo actuates the distal flap for contact-based interaction. A second SG90 servo drives the dispensing metering plate, allowing individual items to be loaded and released through the arm channel. An LED-LDR pair provides feedback at the metering stage to verify successful loading and dispensing. Camera feedback supports target alignment and manipulator adjustment during execution, while the configurable actuation interface allows the platform to accommodate alternative end-effectors and interaction strategies.

%The shared manipulator supports both example tasks using a common target-alignment procedure, thereby reducing hardware complexity and demonstrating the adaptability of the mechanical interface.For the plant-toppling task, the arm is maintained in an approximately vertical configuration and positioned above the selected target. The distal flap then rotates through the target to displace it from its initial position. The targets consist of 3D-printed plant surrogates secured to the testbed using matched magnetic pairs. This arrangement provides repeatable resistance and allows the targets to be restored between trials.

The seed-loading mechanism consists of a four-tube storage tower and a 7-mm-thick rotary metering plate positioned beneath the tubes. Each tube stores multiple seed surrogates, while the metering plate contains a single cavity sized to accept one surrogate. A second SG90 micro servo rotates the plate between the storage tubes and a fixed opening connected to the manipulator channel. During loading, the cavity is first aligned beneath a selected storage tube to receive one surrogate and then rotated toward the channel opening. An LED and a light-dependent resistor monitor the cavity to confirm the presence of a seed surrogate. If no surrogate is detected, the plate advances to another storage tube and repeats the loading procedure. The single-object cavity provides mechanical singulation, while the optical sensor confirms successful loading before manipulator actuation. The primary sensing and actuation components and their functional roles are summarized in Table~\ref{tab:sensing-actuation}. These devices are coordinated through the hierarchical computing and control architecture described in the following subsection.

\begin{table}[H]
\centering
\caption{Primary sensing and actuation components of the Agibot platform.}
\label{tab:sensing-actuation}
\footnotesize
\setlength{\tabcolsep}{3pt}
\renewcommand{\arraystretch}{1.15}

\begin{tabularx}{\columnwidth}{p{0.25\columnwidth}p{0.39\columnwidth}X}
\toprule
\textbf{Category} & \textbf{Component} & \textbf{Function} \\
\midrule
\multirow{2}{*}{Vision} & Logitech C270 (upper) & Navigation and task validation \\
 & Logitech C270 (lower) & Task detection \\
\midrule
Proximity & VL53L1X ToF sensor ($\times$7) & Obstacle and target ranging \\
\midrule
Inertial & LSM6DSOx + LIS3MDL IMU & Orientation and heading \\
\midrule
\multirow{2}{*}{Proprioceptive} & AS5048A magnetic encoder ($\times$4) & Wheel rotation feedback \\
 & STS3215 built-in encoder ($\times$9) & Joint angle feedback \\
\midrule
Verification & LED + LDR pair & Seed-load confirmation \\
\midrule
Locomotion & 12 V DC gear motor ($\times$4) & Wheel drive \\
\midrule
Locomotion \& Manipulation & STS3215 servo ($\times$9) & Steering and arm actuation \\
\midrule
Manipulation & SG90 micro servo ($\times$2) & Flap and seed metering \\
\bottomrule
\end{tabularx}
\end{table}

\subsection{Computing and control architecture}
\label{sec:electronics}

\begin{figure*}
  \centering
  \includegraphics[width=0.77\textwidth]{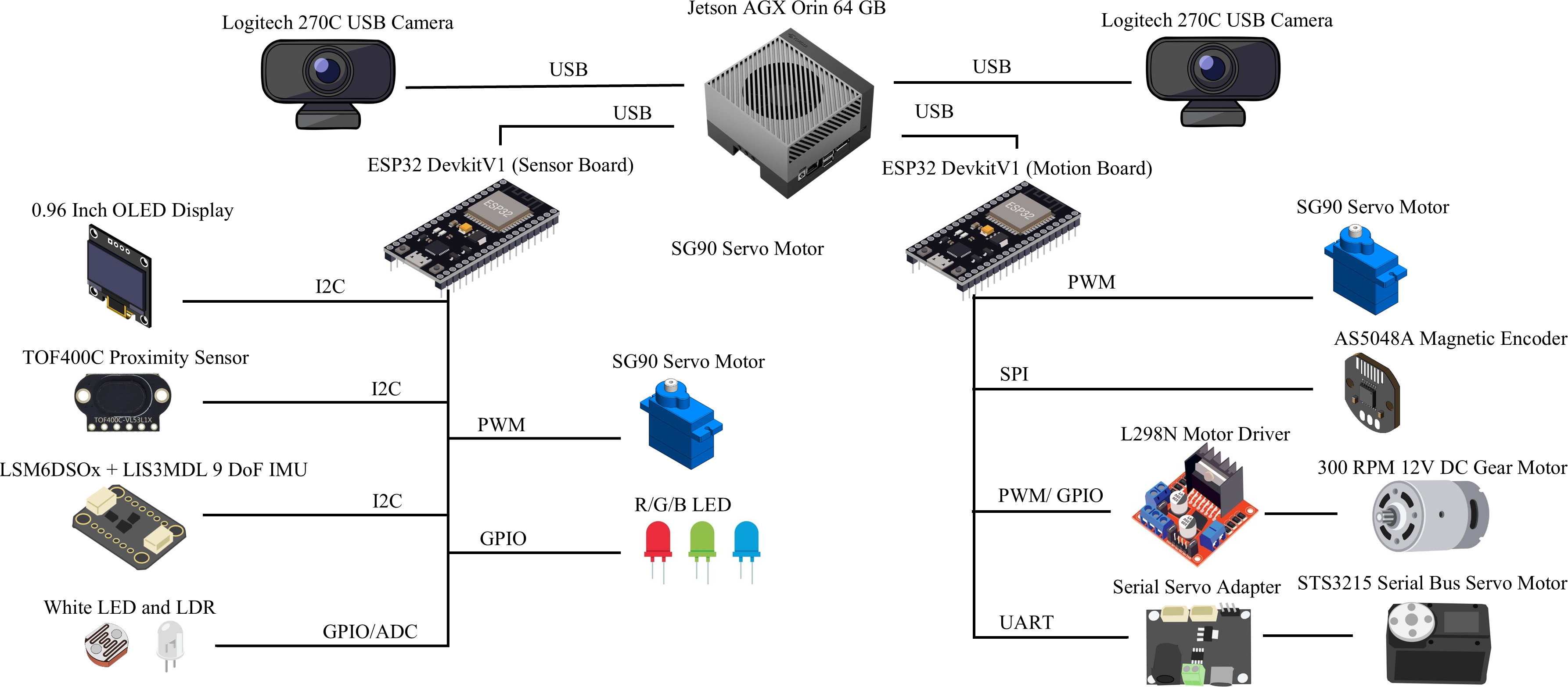}
  \caption{Hierarchical electronics architecture of Agibot. A Jetson AGX Orin performs high-level computation and communicates with two ESP32 microcontrollers, which interface with the platform sensors and actuators through USB, I\textsuperscript{2}C, SPI, UART, PWM, GPIO, and ADC connections.}
  \label{fig:electronics}
\end{figure*}

The platform uses a hierarchical single-master, multi-slave computing architecture as shown in Fig.~\ref{fig:electronics}. A swappable high-level computer performs perception, state estimation, planning, and supervisory control, while two ESP32 microcontrollers manage low-level sensing and actuation through USB serial communication. This separation isolates computationally intensive autonomy algorithms from time-sensitive hardware operations and allows the high-level computer to be replaced without redesigning the embedded control architecture~\cite{godoy2010design,shamshiri2024modular}.

An NVIDIA Jetson AGX Orin was used as the high-level computer in the evaluated configuration. The Jetson processes the two Logitech C270 camera streams received through USB, executes the vision models, estimates platform and target states, and generates navigation and manipulation commands. The computing interface can also support alternative devices, including Raspberry Pi 5, Raspberry Pi 5 with Hailo-8 acceleration, Jetson Orin Nano, and other single-board computers, provided that the required communication and computational interfaces are available. A cross-platform comparison of these computing devices is provided in Appendix~\ref{app:perception-evaluation}.

The two ESP32 microcontrollers are integrated on a single perforated prototyping board with the required communication, power-conditioning, and interface circuitry. The first, designated as the \textit{Sensor Board}, interfaces with the 0.96-in. OLED display, seven TOF400C modules, the IMU, the LED-LDR verification pair, an SG90 servo, and the RGB status LEDs. The TOF400C sensors share an I2C bus, with their XSHUT lines used during startup to assign unique addresses; the OLED and IMU are also connected to the same I2C bus. This shared bus architecture allows additional compatible sensors to be incorporated with minimal wiring changes. The second ESP32, designated as the \textit{Motion Board}, reads the four AS5048A wheel encoders through SPI, drives the 12-V DC motors through the motor drivers, controls an SG90 servo, and sends position commands to the STS3215 while receiving feedback on the position for steering, manipulator, and camera-module servos through a Serial Bus Adapter. The SPI interface supports additional compatible encoder devices, while the addressable STS3215 serial bus allows further servos to be added for alternative manipulators or actuation modules. This division separates sensing and status-related functions from locomotion and actuator control while retaining straightforward expansion of both sensing and actuation hardware.

The high-level computer communicates with both embedded controllers (Fig.~\ref{fig:electronics}) through USB serial connections. It transmits operating-mode selections, wheel-motion commands, and servo-position targets, while the embedded controllers return sensor measurements, actuator states, and command-completion acknowledgments. These acknowledgments allow the supervisory controller to verify the completion of steering and manipulator motions before initiating dependent operations. 

The motion controller converts high-level navigation commands and estimated alignment errors into coordinated drive and steering commands for the four swerve modules. During row traversal, lateral and heading corrections are generated concurrently while maintaining the nominal strafing motion, taking advantage of the platform's independent wheel steering and drive control. The resulting wheel-level commands are updated during each control cycle by the embedded motion controller. The detailed control formulation and implementation are presented in Section~\ref{sec:autonomous-navigation}.

The software architecture mirrors the hardware hierarchy. perception, state estimation, planning, and supervisory decision-making are implemented in Python on the Jetson, while sensor acquisition, motor control, servo communication, and command acknowledgment are implemented in the ESP32 firmware. The high-level software determines the required platform behavior and issues task-level commands. The embedded controllers translate these commands into low-level actuator operations and return sensor and execution-status information, thereby supporting coordinated and verifiable operation across the platform.

\section{Autonomy Methodology} 
\label{sec:autonomy-methods}

\begin{figure*}[!t]
  \centering
  \includegraphics[width=0.85\textwidth]{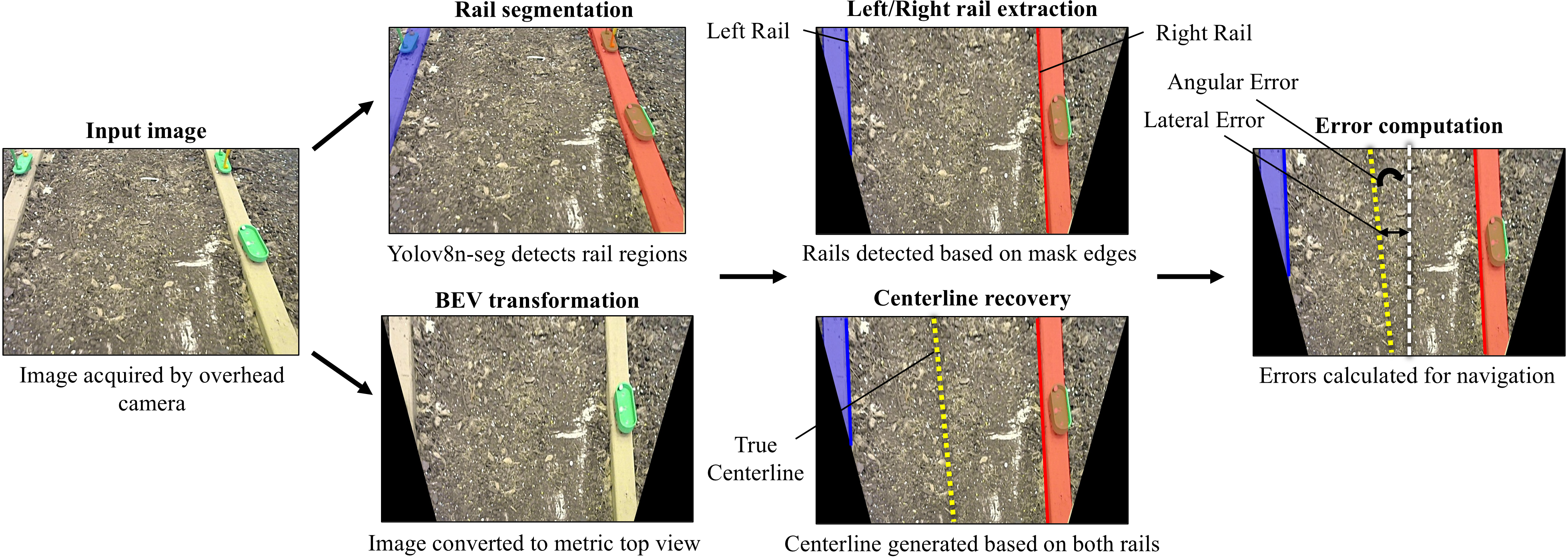}
  \caption{RailNav lane-centerline detection pipeline: the camera's raw image is segmented into left-rail (blue) and right-rail (red) masks, reprojected into a bird's-eye-view (BEV) metric plane, and reduced to the innermost edges of the two rail masks, shown as corresponding blue and red lines. A lane centerline is then reconstructed from the midpoints between the two rail-edge lines and compared with the camera-frame centerline representing the robot heading to compute lateral and yaw errors.}
  \label{fig:railnav}
\end{figure*}

Agricultural manipulation requires more than generating actuator commands because successful task execution depends on whether the robot reaches the appropriate location, identifies the intended task, aligns with the target, completes the physical interaction, and confirms the resulting environmental change. To support this sequence, the autonomy system integrates perception, navigation, manipulation, and task verification within a closed-loop \textit{sense-plan-act-verify} framework. Sensor observations are used to estimate the platform, path, and target states (\textit{sense}); the supervisory controller then selects the required navigation or manipulation behavior and generates the corresponding commands (\textit{plan}); the selected commands are executed by the embedded controllers and actuators (\textit{act}); and post-action perception determines whether the intended physical outcome has occurred (\textit{verify}). If verification fails, the supervisory controller initiates an appropriate recovery or retry action.

The remainder of this section follows the operational sequence of the autonomy framework. Section~\ref{sec:autonomous-navigation} describes perception and control for navigating the platform to the task region. Section~\ref{sec:target-manipulation}  then presents the shared task-execution workflow shown in Fig.~\ref{fig:wheel-control}, including task detection, target alignment, and the task-specific manipulation procedures for plant toppling and seed dispensing. Finally, Section~\ref{sec:task-verification}  describes the command- and outcome-level feedback mechanisms that close the autonomy loop and determine whether execution proceeds, is adjusted, or is repeated.

\subsection{Autonomous navigation} 
\label{sec:autonomous-navigation}

Autonomous navigation combines rail-based path estimation, swerve-drive motion control, and finite-state behavior planning. Images from the upper camera are processed using an instance-segmentation model (e.g., Yolov8n-seg \cite{yolov8_ultralytics}) to identify the left and right rails defining the navigation corridor, and the raw image is concurrently reprojected into a bird's-eye-view (BEV) metric plane. The detected rail pixels are transformed from image coordinates into this metric bird's-eye-view plane using a planar homography:

\begin{equation}
\begin{bmatrix} x_w \\ y_w \\ 1 \end{bmatrix} \sim H \begin{bmatrix} u \\ v \\ 1 \end{bmatrix}, \qquad
H = \begin{bmatrix} h_{11} & h_{12} & h_{13} \\ h_{21} & h_{22} & h_{23} \\ h_{31} & h_{32} & 1 \end{bmatrix},
\label{eq:homography}   
\end{equation}

where $(u,v)$ denotes an image-pixel location, $(x_w,y_w)$ denotes the corresponding ground-plane coordinate, and $H$ is obtained offline from four image-to-ground correspondences. Connected-component analysis is applied to the masks transformed via \eqref{eq:homography}, and the innermost edges of the detected rails are used to define the navigable corridor. A filtered estimate of rail spacing is maintained during operation. If one rail becomes temporarily unavailable because of glare, debris, or partial occlusion, the missing boundary is reconstructed from the detected rail and the estimated rail spacing. The path centerline is calculated as the midpoint between the two boundaries, from which the lateral and angular navigation errors are determined by comparing the recovered centerline against the robot's heading, completing the detection-to-error pipeline shown in Fig.~\ref{fig:railnav}.

\begin{figure}[H]
  \centering
  \includegraphics[width=0.95\linewidth]{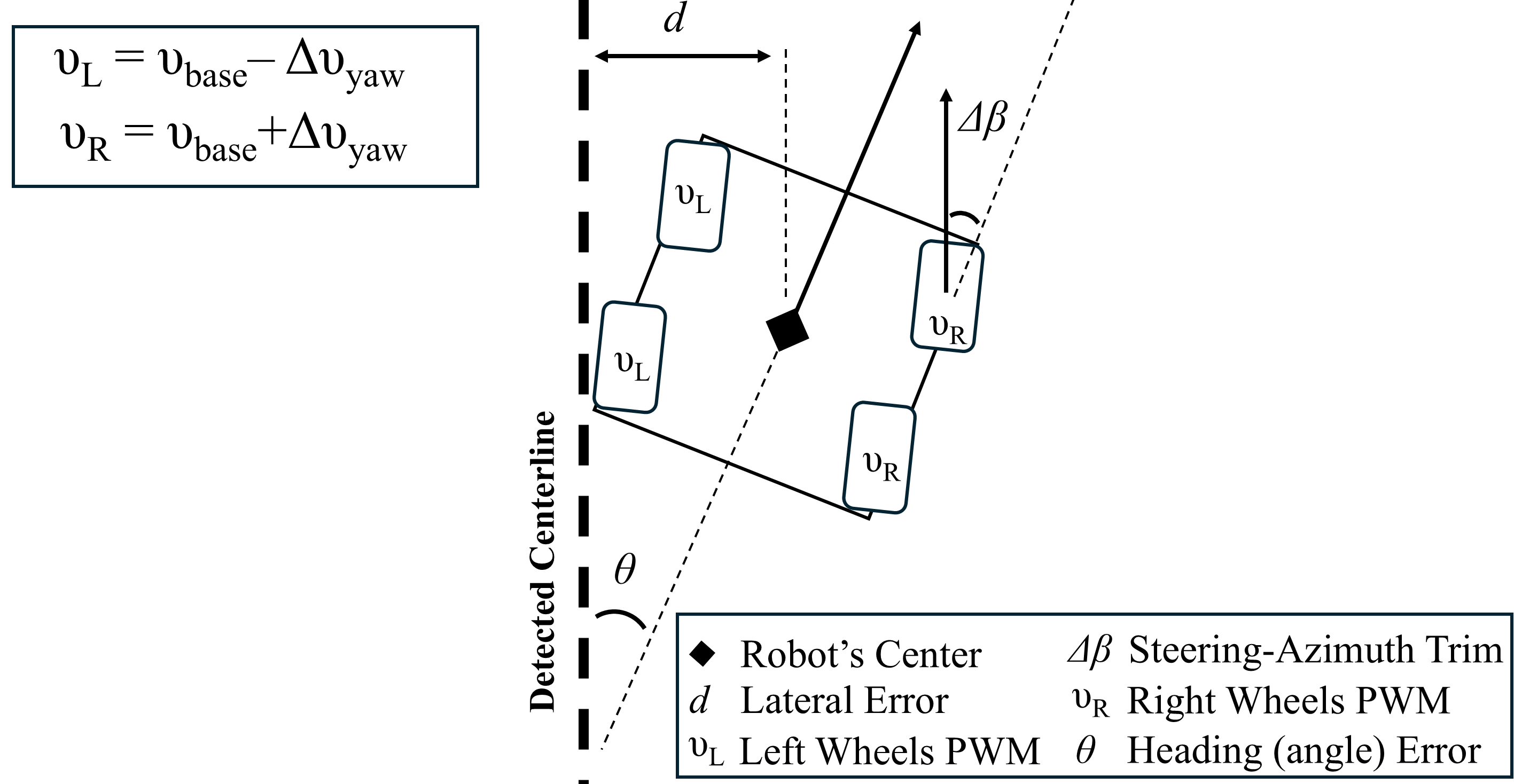}
  \caption{Wheel-level correction geometry during row traversal. The lateral offset and heading deviation from the centerline are exaggerated for
  visualization.}
  \label{fig:wheel-control}
\end{figure}

\begin{figure*}[ht]
  \centering
  \includegraphics[width=0.85\textwidth]{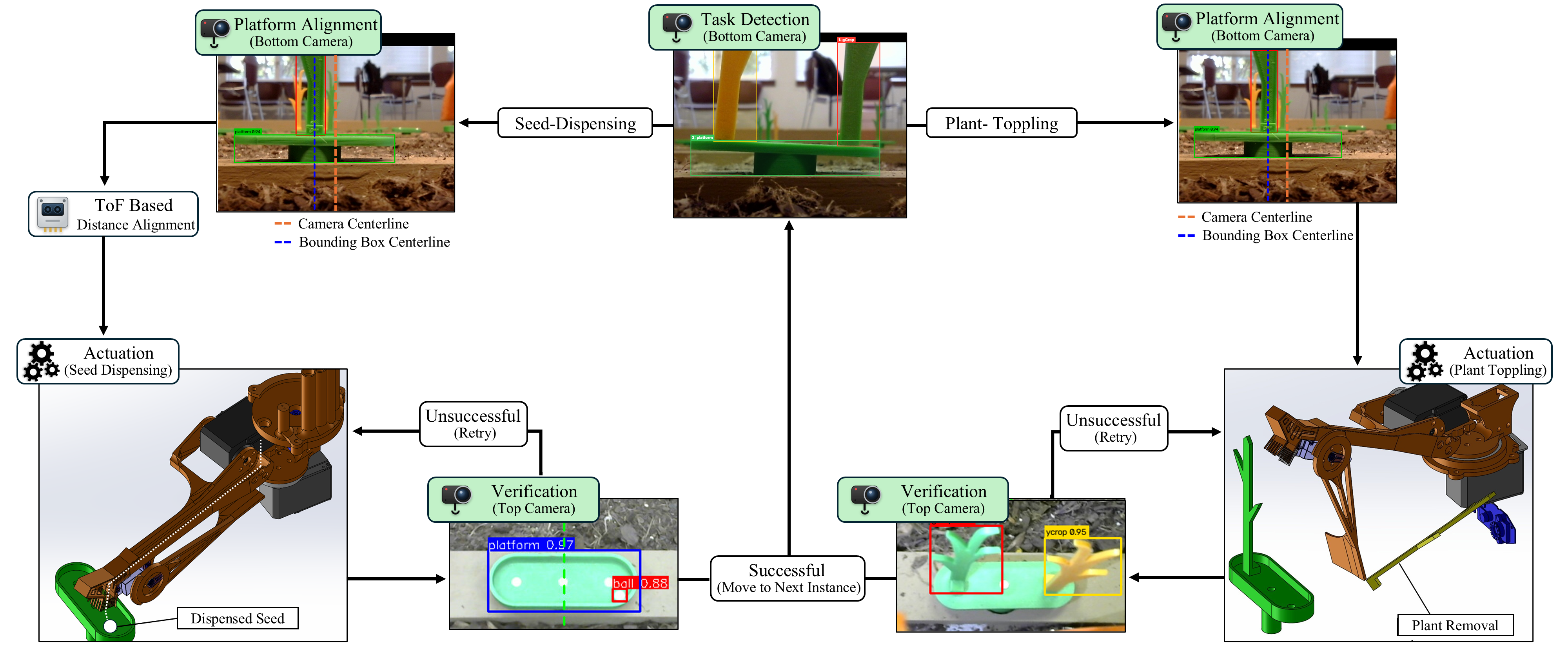}
  \caption{Closed-loop autonomy pipeline for seed-dispensing (left) and plant-toppling (right).
Both begin with bottom-camera platform alignment; seed-dispensing adds ToF-based distance
alignment. Task detection, actuation, and top-camera verification follow, with automatic
retry on failure.}
  \label{fig:closed-loop}
\end{figure*}

During row traversal, the platform moves laterally with the wheels oriented approximately perpendicular to the chassis longitudinal axis. As illustrated in Fig.~\ref{fig:wheel-control}, the controller regulates the lateral error $d$ and heading error $\theta$ relative to the detected centerline. Lateral correction is implemented as a crab-walk-style adjustment by applying a steering-azimuth trim $\Delta\beta$ to all four wheels, introducing a small longitudinal velocity component toward the centerline. Heading correction is generated through a differential wheel-speed term $\Delta v_{yaw}$ between the front and rear wheel pairs; depending on the required yaw direction, one pair is increased above the base strafe speed while the opposite pair remains at the base value, producing a corrective yaw moment without interrupting lateral motion. Both corrections use the same deadbanded and saturating control form: the correction is zero within the corresponding deadband, increases from a minimum nonzero magnitude once the deadband is exceeded, varies linearly with error until the saturation threshold is reached, and remains clamped beyond that threshold, as defined in \eqref{eq:yaw-correction} and \eqref{eq:steering-trim}.
\begin{equation}
\Delta v_{yaw} =
\begin{cases}
0, & |\theta| \leq \theta_{db}, \\[3pt]
\mathrm{sgn}(\theta)\!
\left[
\delta_{min}
+\gamma_{\theta}(\delta_{max}-\delta_{min})
\right],
& |\theta| > \theta_{db},
\end{cases}
\label{eq:yaw-correction}
\end{equation}

\begin{equation}
\Delta\beta =
\begin{cases}
0, & |d| \leq d_{db}, \\[3pt]
\mathrm{sgn}(d)\!
\left[
\beta_{min}
+\gamma_{d}(\beta_{max}-\beta_{min})
\right],
& |d| > d_{db}.
\end{cases}
\label{eq:steering-trim}
\end{equation}

where
$\gamma_{\theta} =
[\mathrm{clip}(|\theta|,\theta_{db},\theta_{sat})-\theta_{db}]
/(\theta_{sat}-\theta_{db})$
and
$\gamma_{d} =
[\mathrm{clip}(|d|,d_{db},d_{sat})-d_{db}]
/(d_{sat}-d_{db})$.
Here, $\theta_{db}$ and $d_{db}$ are the heading and lateral error deadbands,
$\theta_{sat}$ and $d_{sat}$ are the corresponding saturation limits, and
$\delta_{min}/\delta_{max}$ and $\beta_{min}/\beta_{max}$ are the minimum and
maximum correction magnitudes for the yaw and steering-azimuth terms,
respectively.

The two corrective terms are applied simultaneously during continuous strafing, allowing cross-track and heading errors to be regulated without interrupting row traversal. Continuous control is used for row following, whereas discrete motion commands are used for fine alignment near row transitions and manipulation targets. The commanded travel speed is progressively reduced near the end of a row while both corrective terms remain active. Once the platform reaches the designated task region, control transitions from continuous row following to the target-detection, alignment, and manipulation sequence described in Section~\ref{sec:target-manipulation}.

\subsection{Deep learning-based target detection and manipulation} 
\label{sec:target-manipulation}
After navigation brings the platform to a manipulation region, the autonomy system must determine the required task, align the robot with the corresponding target, and coordinate the subsequent physical interaction. The high-level execution sequence for both manipulation tasks is shown in Fig.~\ref{fig:closed-loop}. In both cases, the bottom camera first provides visual information for task detection and platform alignment. A deep learning-based detector localizes and classifies task-relevant regions, including the platform and target-crop regions, and the resulting detections are used to distinguish between plant-toppling and seed-dispensing tasks. The detected target geometry then provides feedback for fine platform alignment; seed dispensing additionally uses ToF measurements for distance alignment before manipulation begins. Although the two tasks require different actuation sequences, both follow the same overall progression of task detection, alignment, actuation, and post-action verification.

For plant toppling, the platform uses image-based position feedback from the bottom camera to laterally align the selected target with the manipulator. Once the target enters the predefined alignment region, the manipulator is positioned above it and the distal flap is actuated to displace the target. For seed dispensing, the robot first performs image-based alignment with the target container and then approaches to a predefined distance using ToF feedback. The metering mechanism transfers one seed surrogate from the storage tower into the manipulator channel, where seed presence is confirmed before the arm is positioned toward the target and the surrogate is released under gravity. These two tasks therefore share the same high-level autonomy structure while differing in their target-alignment requirements and manipulation actions.

\subsection{Closed-loop validation} 
\label{sec:task-verification} 
The final stage of the task-execution workflow shown in Fig.~\ref{fig:closed-loop} closes the autonomy loop by evaluating execution at two levels: command completion and physical task outcome. At the command-execution level, wheel and servo feedback confirms that the required steering and manipulator positions have been reached before dependent actions are initiated. For seed dispensing, the seed-loading sensor additionally confirms that a surrogate has entered the manipulator channel before the release sequence proceeds.

At the physical-outcome level, the top camera inspects the target region after manipulation using a separately trained deep learning-based verification model. For seed dispensing, the system determines whether the released seed surrogate is present within the target container; for plant toppling, it determines whether the selected target remains in its original upright position. A verification result is accepted only when the same outcome is observed across multiple consecutive frames, reducing sensitivity to intermittent detections. If the intended outcome is not confirmed, the supervisory controller follows the retry path shown in Fig.~\ref{fig:closed-loop} and repeats the relevant actuation step; otherwise, the platform proceeds to the next task instance.

Additional recovery mechanisms operate during fine alignment. A stuck-wheel monitor compares commanded movement with encoder feedback and incrementally increases the drive command, within predefined limits, when insufficient wheel motion is detected. Continuous longitudinal motion is also bounded by a fixed duration, after which the controller transitions to distance-limited alignment. Together, these mechanisms close the loop between commanded actions and verified physical outcomes, ensuring that task completion is determined from the resulting environmental change rather than from actuator commands alone.

\section{Experimental Evaluation}
\label{sec:experimental-evaluation}

\subsection{Experimental setup} 
\label{sec:experimental-setup}
The platform was evaluated at both the subsystem and integrated-system levels on a representative, soil-configurable agricultural testbed containing parallel navigation rails, row-transition areas, manipulation targets, and an asymmetric course layout requiring autonomous recovery. The rails defined the navigation corridor traversed by the platform, while 3D-printed plant representations and target containers, visible during the closed-loop manipulation sequences, were used to evaluate plant toppling and seed dispensing, respectively. Further details on controller-parameter tuning and subsystem-level evaluations are provided in Appendices~\ref{app:perception-evaluation}-\ref{app:manipulation-performance}.

Subsystem experiments assessed perception, navigation, and manipulation performance, while integrated trials evaluated the complete \textit{sense-plan-act-verify} workflow under varying lighting and surface conditions. The complete system was evaluated under five environmental conditions: low illumination, high illumination, no soil, a thin soil layer, and a heavy soil layer. The illumination experiments were conducted with an intermediate soil layer, while the soil experiments were performed under intermediate illumination. Five repeated trials were conducted for each experimental configuration.

Navigation performance was evaluated using lateral root mean square error (RMSE), angular RMSE, maximum lateral error, course completion, and completion time. Manipulation performance was evaluated using the numbers of attempted and successful actions, misplaced objects, and execution time. Integrated-system performance was evaluated using course completion, successful actuations, and total execution time. A trial was considered successful when the platform reached the end of the course after completing the prescribed navigation and manipulation sequence. Detailed perception-model comparisons, controller-parameter studies, and trial-level results are provided in the Appendix.

\begin{table}[H] 
\centering 
\caption{RailNav performance on 68 held-out images. Bold values indicate the better result for each metric. Both models achieved a valid-centerline rate of 97.06\%. } 
\label{tab:railnav-quality} \footnotesize \setlength{\tabcolsep}{6pt} \renewcommand{\arraystretch}{1.08} 

\begin{tabular}{lccc} 
\toprule 
\textbf{Metric} & \textbf{YOLOv8n-seg} & \textbf{YOLO11n-seg} \\ \midrule 
Mask $\mathrm{mAP}_{50:95}$ & 0.950 & \textbf{0.958} \\ 
Valid centerlines & 66/68 & 66/68 \\ 
Near-field MAE (px) & \textbf{2.471} & 3.051 \\ 
Far-field MAE (px) & 0.756 & \textbf{0.731} \\ 
Path-angle MAE ($^\circ$) & \textbf{0.795} & 0.893 \\ 
Mean centerline jitter (px) & 3.110 & \textbf{2.763} \\ \bottomrule \end{tabular}
\end{table}

\subsection{Subsystem performance} 
\label{sec:subsystem-performance}
The RailNav perception pipeline was evaluated using 68 held-out images. Two representative instance-segmentation models, YOLOv8n-seg \cite{yolov8_ultralytics} and YOLO11n-seg \cite{yolo11_ultralytics}, were compared in terms of segmentation
accuracy, centerline-estimation quality, and deployment cost, although the
underlying pipeline is not restricted to these two models and can
accommodate other instance-segmentation architectures. As reported in Table~\ref{tab:railnav-quality}, both models generated a valid navigation centerline for 66 of the 68 images, corresponding to a solution rate of 97.06\%. YOLOv8n-seg achieved a mask $\mathrm{mAP}_{50:95}$ of 0.9497 and mean absolute near-field, far-field, and path-angle errors of 2.471 pixels, 0.756 pixels, and $0.795^{\circ}$, respectively. Although YOLO11n-seg produced slightly higher mask accuracy and lower centerline jitter, YOLOv8n-seg provided lower near-field and path-angle errors.

The computational requirements of the two models were subsequently compared after FP16 TensorRT deployment on the Jetson AGX Orin. As shown in Table~\ref{tab:railnav-deployment}, YOLOv8n-seg required 159.208~ms for the complete model-inference, mask-processing, and bird's-eye-view navigation pipeline, compared with 161.269~ms for YOLO11n-seg. Because the two models provided comparable centerline-estimation performance, YOLOv8n-seg was selected for deployment based on its lower end-to-end latency and lower GPU processing time.

\begin{table}[H] 
\centering 
\caption{Runtime performance of the RailNav models after FP16 TensorRT deployment. Bold values indicate the deployed model. } 
\label{tab:railnav-deployment} 
\footnotesize \setlength{\tabcolsep}{3.5pt} 
\renewcommand{\arraystretch}{1.08}
\begin{tabular}{lccc} 
\toprule 
\textbf{Model} & \makecell{\textbf{GPU time}\\
\textbf{(ms)}} & \makecell{\textbf{BEV/navigation}\\\textbf{(ms)}} & \makecell{\textbf{Total latency}\\\textbf{(ms)}} \\ \midrule YOLOv8n-seg & \textbf{2.167} & \textbf{142.879} & \textbf{159.208} \\ 
YOLO11n-seg & 2.537 & 143.776 & 161.269 \\ \bottomrule 
\end{tabular} 
\end{table}

\begin{table*}[!t]
\centering
\footnotesize
\setlength{\tabcolsep}{9pt}
\caption{Navigation performance under different controller-gain and target-speed settings. Values represent the means of five trials.}
\label{tab:navigation-sweep}
\begin{tabular}{llccccc}
\toprule
\textbf{Gain} & \textbf{Speed} & \textbf{Lateral RMSE} & \textbf{Angular RMSE} & \textbf{Max. lateral error} & \textbf{Completion time} & \makecell{\textbf{Mean course}\\\textbf{completion}} \\
\midrule
Nominal & Low     & 34.58 px & 1.20\textdegree & 294.00 px & 194.09 s & 100\% \\
Nominal & High    & 44.18 px & 1.98\textdegree & 257.00 px & 132.62 s & 95\% \\
\textbf{Nominal} & \textbf{Nominal} & \textbf{42.92 px} & \textbf{1.59\textdegree} & \textbf{228.00 px} & \textbf{141.51 s} & \textbf{100\%} \\
High    & Nominal & 58.75 px & 2.35\textdegree & 289.00 px & 143.85 s & 100\% \\
Low     & Nominal & 31.50 px & 1.15\textdegree & 175.00 px & 139.80 s & 100\% \\
\bottomrule
\end{tabular}
\end{table*}

\begin{table*}[!t] 
\centering 
\caption{Integrated autonomous performance with and without post-action task verification. Values represent the means of five trials, with 26 manipulation opportunities per trial.} 
\label{tab:integrated-performance} 
\footnotesize \setlength{\tabcolsep}{5pt} 
\begin{tabular}{llcccc} \toprule \textbf{Condition} & \textbf{Verification} & \textbf{Successful Trials} & \textbf{Course Completion (\%)} & \textbf{Successful Actions per Trial} & \textbf{Execution Time (s)} \\ \midrule Low light & No & 5/5 & 100 & 25.6 & 526.67 \\ Low light & Yes & 5/5 & 100 & 26.0 & 595.39 \\ \hline  
High light & No & 5/5 & 100 & 25.2 & 526.42 \\ High light & Yes & 5/5 & 100 & 25.8 & 589.25 \\ \hline  
No soil & No & 5/5 & 100 & 25.4 & 484.89 \\ No soil & Yes & 5/5 & 100 & 25.6 & 552.15 \\  \hline  
Thin soil & No & 5/5 & 100 & 24.4 & 522.08 \\ Thin soil & Yes & 5/5 & 100 & 25.6 & 597.52 \\  \hline  
Heavy soil & No & 4/5 & 97 & 25.2 & 572.30 \\ 
Heavy soil & Yes & 5/5 & 100 & 25.8 & 652.30 \\ 
\bottomrule 
\end{tabular} 
\end{table*}

Navigation performance was evaluated using five combinations of controller gain and target speed, with five trials conducted for each combination. As summarized in Table~\ref{tab:navigation-sweep}, the platform achieved mean course completion rates of at least 95\% under all evaluated settings. At the nominal target speed, the low-gain setting produced the lowest lateral RMSE of 31.50 pixels, angular RMSE of $1.15^{\circ}$, and maximum lateral error of 175 pixels. At the nominal gain, increasing the target speed reduced the mean completion time from 194.09~s at low speed to 132.62~s at high speed. However, the high-speed setting increased the lateral and angular RMSE values and reduced mean course completion to 95\%. These findings illustrate the tradeoff between traversal speed and tracking performance. The nominal-gain, nominal-speed setting was retained for the integrated experiments because it provided 100\% mean course completion and stable behavior across repeated trials. Although the low-gain setting produced better aggregate tracking metrics, the nominal setting was selected as a consistent operating point for subsequent system-level evaluation.

The two proof-of-concept manipulation tasks were evaluated independently over five trials. Plant toppling succeeded in all 43 attempted actions, with a mean actuation time of 3.33~s. Seed dispensing succeeded in 52 of 55 attempted actions, corresponding to a success rate of 94.5\% and a mean actuation time of 9.77~s. The three unsuccessful dispensing actions comprised two misplaced releases and one incomplete dispensing event. These results characterize the repeatability of the manipulation tasks before their integration into the complete autonomy workflow.

\subsection{Integrated system performance} 
\label{sec:integrated-performance}
Integrated trials evaluated the complete autonomy workflow with and without post-action visual verification. Five approximately 10-min trials, each containing 26 manipulation opportunities, were conducted per condition, yielding 50 trials and 1,300 opportunities across the five conditions and two verification modes. Illumination was varied with an intermediate soil layer, whereas soil conditions were varied under intermediate illumination. Low-level actuator feedback and safety mechanisms remained active throughout all trials. This repeated-action evaluation was intended to establish proof-of-concept system capability rather than statistical characterization of long-term field reliability.

As summarized in Table~\ref{tab:integrated-performance}, 98\% of all trials reached the end of the course across the low-light, high-light, no-soil, thin-soil, and heavy-soil conditions. The only incomplete trial occurred under heavy soil without verification and traversed approximately 85\% of the course, giving 97\% mean completion for that condition. Across all conditions, successful actions increased from 629/650 (96.8\%) without verification to 644/650 (99.1\%) with verification. Thus, inspection and retry produced 15 additional successful outcomes and reduced unsuccessful actions from 21 to 6 (71.4\%); this is a net recovery benefit because individual verification-trigger counts were not logged separately. The largest improvement occurred under thin soil, from 24.4 to 25.6 successful actions per trial. Verification also increased mean trial duration by 62.83--80.00~s, demonstrating improved task reliability at a measurable execution-time cost.

\begin{figure*}[!t]
  \centering
  \includegraphics[width=0.95\textwidth]{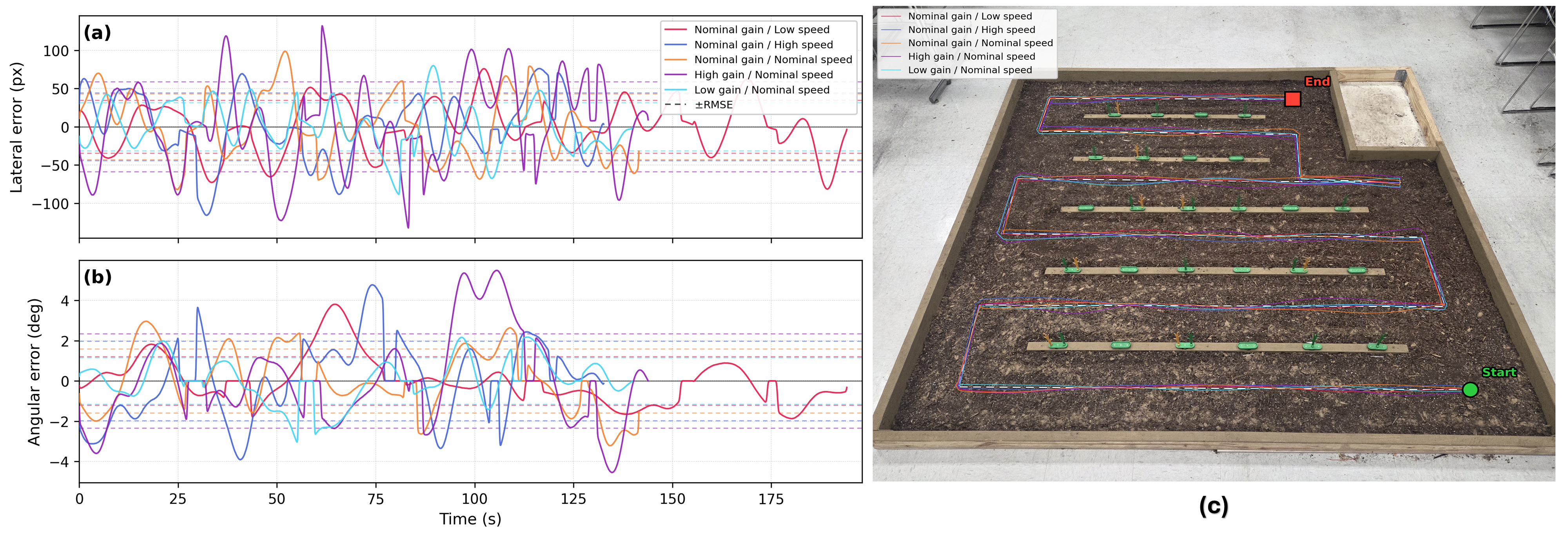}
  \vspace{-5pt}
  \caption{Navigation performance for one representative run at each of the five controller-gain and target-speed settings in Table~\ref{tab:navigation-sweep}: (a) lateral error, (b) angular error, and (c) platform trajectories on the testbed. Dashed lines in (a) and (b) mark each setting's $\pm\mathrm{RMSE}$ in the matching color. Start and end positions are marked in (c).}
  \label{fig:navigation-tracing}
\end{figure*}

To further characterize the navigation behavior underlying these integrated trials, Fig.~\ref{fig:navigation-tracing} shows the lateral-error, angular-error, and trajectory traces for one representative run at each of the five gain and target-speed settings in Table~\ref{tab:navigation-sweep}. Raising the target speed widens and roughens the error band around the dashed $\pm$RMSE lines: the high-speed run (44.18~px lateral RMSE, 95\% completion) exhibits a broader lateral-error distribution than the nominal-speed run (42.92~px, 100\%). Lowering the gain has the opposite effect, producing the tightest of the five traces (31.50~px lateral RMSE, $1.15^{\circ}$ angular RMSE, 175~px maximum lateral error), compared with 42.92~px, $1.59^{\circ}$, and 228~px, respectively, at the nominal gain. As explained in Section~\ref{sec:subsystem-performance}, the nominal-gain, nominal-speed setting was nonetheless retained for the integrated experiments because its behavior was more consistent across the five repeated trials than the low-gain setting.

Overall, the experiments show that Agibot supports integrated perception, navigation, manipulation, and post-action verification under variable environmental conditions. The tuning results show that inappropriate values increase tracking error, overshoot, or settling time. Vision performance also depends on lighting and soil appearance and may require adjustment of mask and threshold parameters. Because evaluation used a representative agricultural testbed rather than long-term field operation, the results establish platform and autonomy-workflow capability but not field-scale reliability; these sensitivities also provide accessible research and instructional variables for controller and perception development.

\section{Conclusion}
\label{sec:conclusion}

This paper presented Agibot, an open-source, modular, swerve-drive platform for agricultural robotics research and education. It integrates perception, control, planning, manipulation, and closed-loop verification in one autonomy architecture. Trials in a representative agricultural testbed across varying lighting, soil, and row-geometry conditions, including row switching and obstruction recovery, demonstrated row-crop navigation, plant toppling, and seed dispensing. Closed-loop verification improved actuation reliability at the cost of additional execution time. A browser-based dashboard and onboard OLED/LED indicators give operators real-time feedback both remotely and locally. 
% The experiments demonstrate its research-testbed capabilities; its educational effectiveness remains to be assessed through planned classroom deployment.
%
Future work will evaluate the platform under real-field conditions, with additional sensors, end-effectors, and agricultural operations. It will also be integrated into robotics courses through lab modules covering perception, embedded control, navigation, and manipulation. Student engagement in platform development and validation will help assess its effectiveness for hands-on, interdisciplinary robotics education.

\vspace{-8pt}
\section*{Acknowledgment}
The authors would like to thank the support from the Mississippi Agricultural and Forestry Experiment Station (MAFES) and the U.S. Department of Agriculture (USDA) Hatch Multi-State project under accession number MIS-300023. Any opinions, findings, conclusions, or recommendations expressed in this publication are those of the authors and should not be construed to represent any views of the funding agency and/or the institutions that the authors are affiliated with.

\bibliographystyle{IEEEtran}
\bibliography{references}

\begin{figure*}[!t]
  \centering
  \includegraphics[width= 0.92 \textwidth]{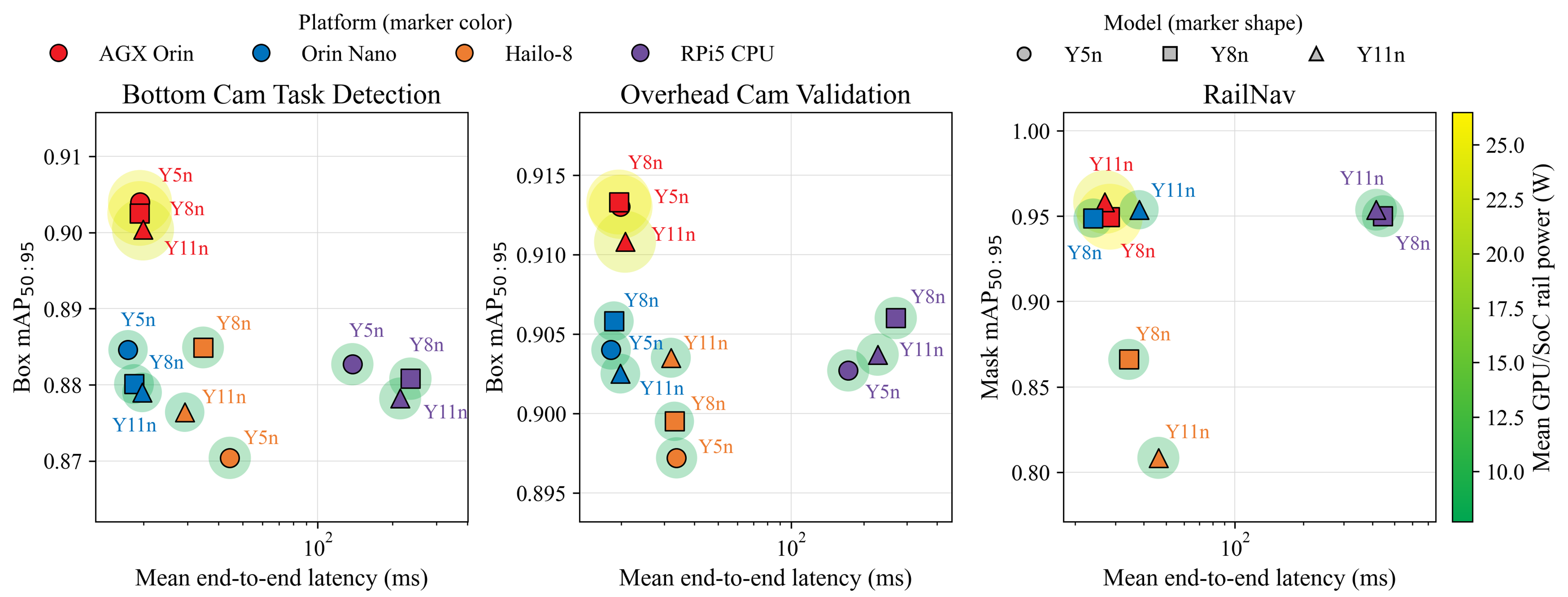}
  \caption{Accuracy vs.\ deployment cost across four platforms, i.e., AGX Orin, Orin Nano, Hailo-8, and RPi5 CPU,
  for task-detection, task-validation, and RailNav segmentation (Box mAP$_{50:95}$; Mask mAP$_{50:95}$ for
  RailNav), each at FP16 precision except Hailo-8 (INT8-only). Marker shape denotes model variant, color denotes
  platform, and the translucent halo encodes mean rail power draw. Accuracy is
  largely platform-independent at matched precision -- latency and power separate the platforms.}
  \label{fig:accuracy-performance}
\end{figure*}

\newpage
\section*{APPENDIX}
\appendices
\section{Cross-Platform Perception Evaluation}
\label{app:perception-evaluation}

The three deployed perception models, i.e., task detection, task validation, and the RailNav segmentation model, were additionally benchmarked in isolation on three lower-cost edge platforms: an NVIDIA Jetson Orin Nano, a Raspberry Pi 5 with a Hailo-8 accelerator, and a Raspberry Pi 5 running the same models on CPU alone (NCNN). Fig.~\ref{fig:accuracy-performance} reports accuracy, latency, and power for all three tasks across the four platforms, each at FP16 precision except the Hailo-8, whose accelerator supports only INT8. Across platforms, the two detection tasks showed relatively small accuracy variation (task detection: 0.870-0.904 mAP$_{50:95}$; task validation: 0.895-0.914 mAP$_{50:95}$), whereas RailNav exhibited substantially larger variation (0.808-0.958 mask mAP$_{50:95}$). Thus, detection accuracy was comparatively stable across platforms, while RailNav segmentation performance was more sensitive to the deployment platform and inference backend. Latency and power draw do: the Orin Nano and Hailo-8 match or approach the AGX Orin's per-model latency (17-46~ms) at roughly a third of its power draw (7.7-9.4~W vs.\ 23.2-26.5~W), while the CPU-only Raspberry Pi trades that same low power for latency an order of magnitude higher (138-445~ms).

Two considerations justify deploying YOLOv8n(-seg) on the Jetson AGX Orin for this study's testbed evaluation despite the cheaper platforms' favorable isolated-model results. First, this benchmark evaluates each model in isolation, whereas the robot's control loop runs task detection, task validation, and RailNav concurrently; only the AGX Orin's compute headroom has been validated for sustaining all three models together at the frame rate the control loop requires, so the lower-power platforms' single-model numbers here do not by themselves confirm they could replace it. Second, applying the same lowest-latency selection rule to task detection and task validation on the AGX Orin, YOLOv8n gives the lowest end-to-end latency for both tasks (19.292~ms and 19.514~ms) at accuracy within the observed platform-to-platform spread, making it the consistent choice for the two detection models. For RailNav specifically, YOLO11n-seg shows marginally lower model-only latency and higher mask accuracy in this isolated cross-platform benchmark (26.983~ms, 0.9583) than YOLOv8n-seg (28.366~ms, 0.9497). However, model-only latency does not represent the deployed navigation pipeline, which also includes bird's-eye-view transformation and navigation post-processing. Under the complete deployed pipeline, YOLOv8n-seg produced the lower end-to-end latency (159.208~ms vs.\ 161.269~ms, Table~\ref{tab:railnav-deployment}). Because model selection for the integrated robot trials was based on end-to-end navigation latency rather than isolated inference latency, YOLOv8n-seg was retained for deployment.

\section{Controller Parameter Tuning}
\label{app:controller-tuning}

\begin{figure*}[!t]
  \centering
  \includegraphics[width=0.8\textwidth]{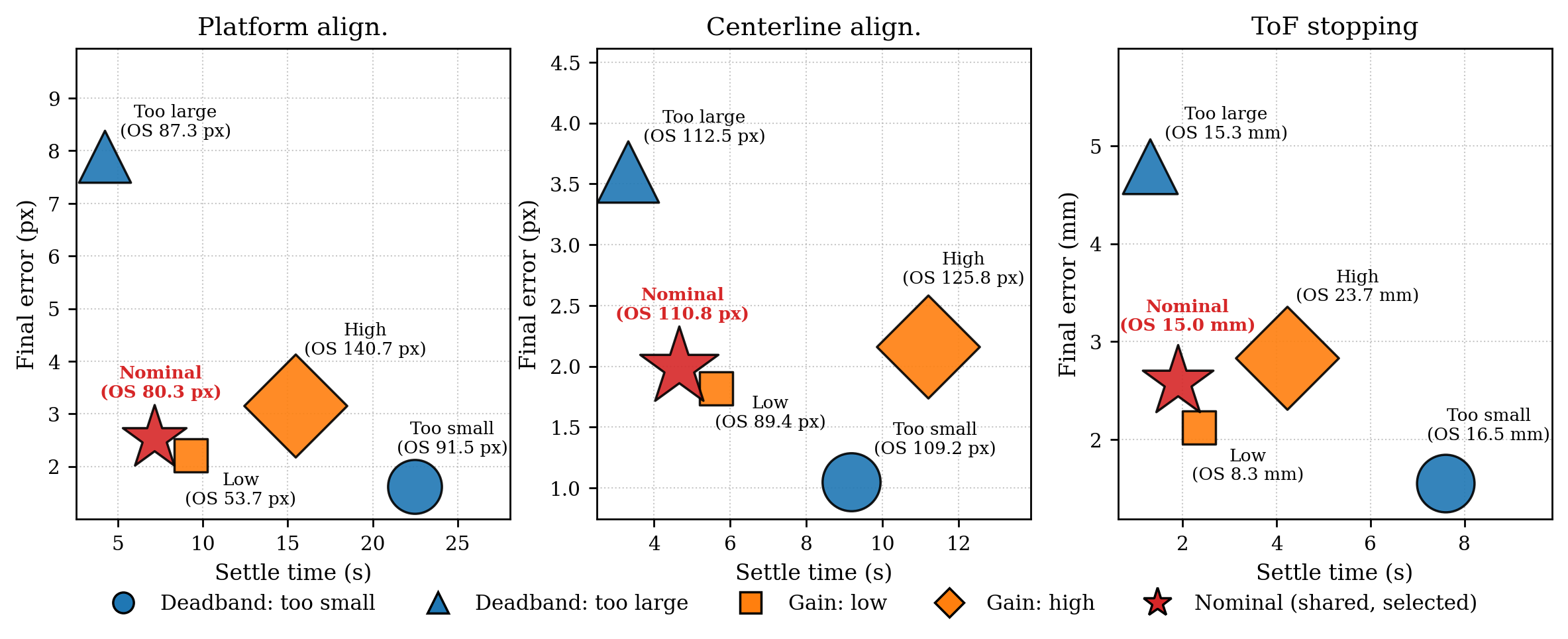}
  \caption{Deadband and gain sweeps combined, across all three tuning targets. Marker size scales with
  overshoot. The nominal setting (red star) gives the shortest or near-shortest settling time in both sweeps
  while conceding little on final error and overshoot.}
  \label{fig:tuning-comparison}
\end{figure*}

The effects of controller deadband and gain were characterized for platform alignment, centerline alignment, and time-of-flight-based stopping. Each parameter was swept one at a time, with the other held at its nominal value, and every setting was evaluated over five repeated trials. Final alignment error, maximum overshoot, and settling time were used as the performance metrics. Fig.~\ref{fig:tuning-comparison} summarizes both sweeps across all three controller targets.

Viewed together, the two sweeps show a consistent pattern across all three targets. Final error and overshoot both improve monotonically as the gain is lowered and as the deadband is narrowed, so the most accurate and least oscillatory setting is always the lowest gain or the smallest deadband. Settling time works against this. For the deadband, it trades off directly against final error, falling monotonically as the band widens. For the gain, it is non-monotonic and reaches a minimum at an intermediate value, since a low gain corrects slowly while a high gain overshoots and must reverse. The nominal setting sits at or near this minimum while conceding only a small amount of final error and overshoot relative to the lowest-gain or narrowest-deadband case. This balance holds across all three targets, which is why the nominal setting was retained as the default operating point; the target-specific trade-offs are detailed next.

\subsection{Deadband sensitivity}
\label{app:deadband-sensitivity}

Deadband width trades settling time against steady-state accuracy, as summarized in Table~\ref{tab:deadband-sweep}. As the band widens, final error grows and settling time falls monotonically across all three targets. For platform alignment, the nominal band cut settling time from 22.48 to 7.15 s relative to the smallest band, at a final error of 2.54 px; the largest band reached 4.21 s but raised final error to 7.89 px. Centerline alignment behaved the same way. The governing objective is task-dependent. For the two alignment targets, fast convergence matters most provided final error stays within tolerance, which favors the nominal band over the smallest. For time-of-flight stopping, overshoot matters most because it carries the platform past its stopping point, and the nominal band gave the lowest overshoot, 14.96 mm, at 1.89 s settling. A small deadband lengthens settling because the controller keeps reacting to errors it can barely resolve; a wide deadband settles quickly but accepts more steady-state error.

\begin{table}[H]
\centering
\footnotesize
\setlength{\tabcolsep}{4pt}
\caption{Effect of deadband width on platform alignment, centerline alignment, and ToF-based stopping. Bold
values indicate the best result within each controller target and metric.}
\label{tab:deadband-sweep}
\begin{tabular}{llccc}
\toprule
\textbf{Target} & \textbf{Setting} & \textbf{Final error} & \textbf{Overshoot} & \textbf{Settle time} \\
\midrule
\multirow{3}{*}{Platform align.}
 & Too small & \textbf{1.62 px} & 91.52 px & 22.48 s \\
 & Nominal   & 2.54 px & \textbf{80.28 px} & 7.15 s \\
 & Too large & 7.89 px & 87.33 px & \textbf{4.21 s} \\
\midrule
\multirow{3}{*}{Centerline align.}
 & Too small & \textbf{1.05 px} & \textbf{109.17 px} & 9.18 s \\
 & Nominal   & 1.99 px & 110.78 px & 4.66 s \\
 & Too large & 3.60 px & 112.52 px & \textbf{3.31 s} \\
\midrule
\multirow{3}{*}{ToF stopping}
 & Too small & \textbf{1.55 mm} & 16.47 mm & 7.60 s \\
 & Nominal   & 2.59 mm & \textbf{14.96 mm} & 1.89 s \\
 & Too large & 4.79 mm & 15.32 mm & \textbf{1.30 s} \\
\bottomrule
\end{tabular}
\end{table}

\subsection{Gain sensitivity}
\label{app:gain-sensitivity}

Final error and overshoot both fell monotonically as the gain was lowered, across all three controllers, as summarized in Table~\ref{tab:gain-sweep}. Settling time did not: it was minimized at the nominal gain for every target and rose at both the low and high settings, because a low gain corrects slowly while a high gain overshoots and must reverse. A high gain also inflated overshoot across all three targets; for platform alignment, settling time rose from 7.15 to 15.47 s and overshoot from 80.28 to 140.74 pixels between the nominal and high settings. A low gain minimized final error and overshoot but converged more slowly. The nominal gain was therefore retained as the default because it gave the fastest convergence without the overshoot growth seen at the high gain. The low-gain setting may be preferable when minimizing overshoot or final error is the primary control objective and slower convergence is acceptable.

\begin{table}[H]
\centering
\footnotesize
\setlength{\tabcolsep}{4pt}
\caption{Effect of controller gain on platform alignment, centerline alignment, and ToF-based stopping. Bold
values indicate the best result within each controller target and metric.}
\label{tab:gain-sweep}

\begin{tabular}{llccc}
\toprule
\textbf{Target} & \textbf{Gain} & \textbf{Final error} & \textbf{Overshoot} & \textbf{Settle time} \\
\midrule
\multirow{3}{*}{Platform align.}
 & Low     & \textbf{2.21 px} & \textbf{53.67 px} & 9.29 s \\
 & Nominal & 2.54 px & 80.28 px & \textbf{7.15 s} \\
 & High    & 3.15 px & 140.74 px & 15.47 s \\
\midrule
\multirow{3}{*}{Centerline align.}
 & Low     & \textbf{1.82 px} & \textbf{89.43 px} & 5.63 s \\
 & Nominal & 1.99 px & 110.78 px & \textbf{4.66 s} \\
 & High    & 2.16 px & 125.75 px & 11.20 s \\
\midrule
\multirow{3}{*}{ToF stopping}
 & Low     & \textbf{2.12 mm} & \textbf{8.28 mm} & 2.35 s \\
 & Nominal & 2.59 mm & 14.96 mm & \textbf{1.89 s} \\
 & High    & 2.83 mm & 23.71 mm & 4.22 s \\
\bottomrule
\end{tabular}
\end{table}

\section{Operator Feedback Interfaces}
\label{app:operator-feedback}

\begin{figure*}[!ht]
  \centering
 \includegraphics[width=0.9\textwidth]{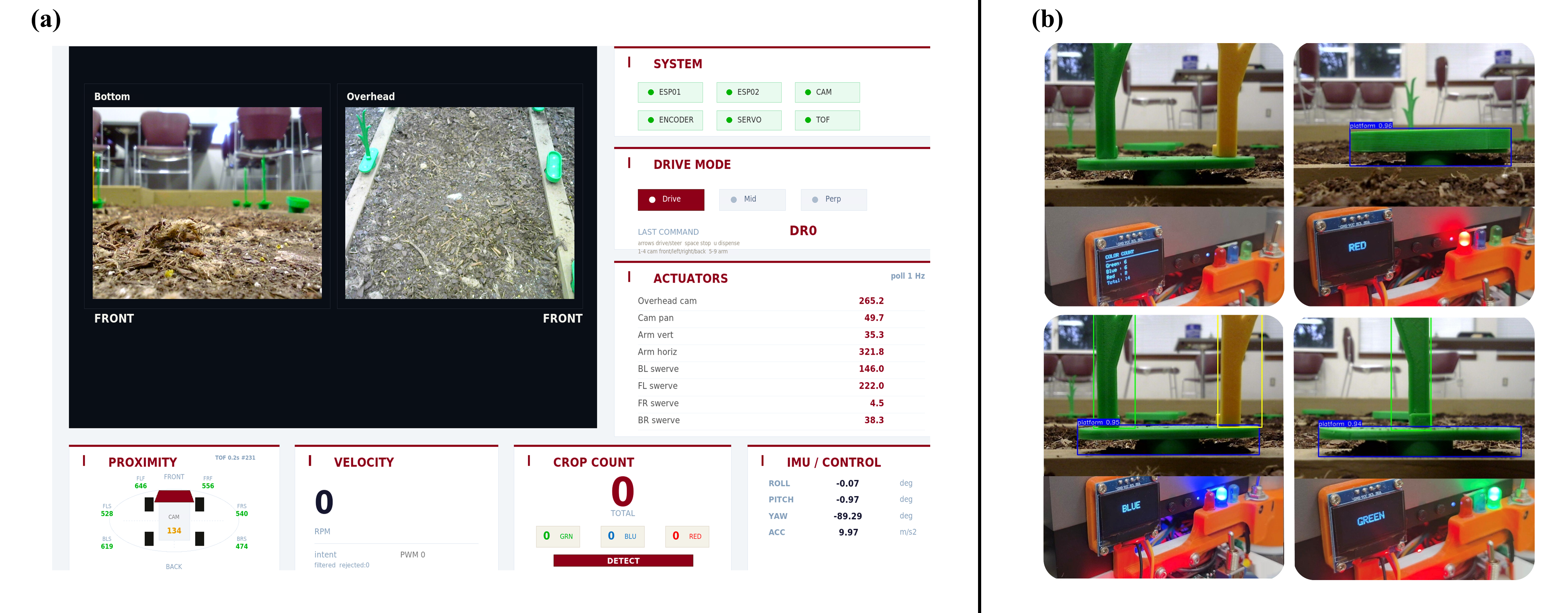}
  \caption{Operator feedback interfaces. (a) Browser-based operator dashboard displaying live camera feeds,
  system status, drive-mode control, actuator positions, proximity measurements, velocity, crop count, and
  IMU/control information. (b) Onboard local feedback: representative OLED and tri-color LED
  readouts for the detected platform states.}
  \label{fig:dashboard}
\end{figure*}

A browser-based operator dashboard, shown in Fig.~\ref{fig:dashboard}(a), provides the primary real-time supervisory view during operation. The interface displays the two camera streams, drive mode, actuator positions, time-of-flight measurements, platform velocity, crop count, inertial measurements, and controller state, along with commands for selecting drive modes and monitoring individual system components. Because the interface operates through a standard web browser, it can be accessed from multiple computing devices without installing a platform-specific application, supporting remote monitoring, debugging, and simultaneous observation by multiple users.

For situations where a connected display is inconvenient, such as observing a testbed trial without a paired laptop or phone, the platform also reports its state locally through an onboard 0.96-in.\ OLED display and a tri-color LED, shown in Fig.~\ref{fig:dashboard}(b). Once the task-detection model classifies the target configuration in front of the robot, the resulting case, coded R, B, or G by which combination of platform, green-crop, and yellow-crop detections is present, is shown on the OLED and mirrored on the tri-color LED, while the OLED simultaneously accumulates a running per-color tally. This lightweight readout lets an observer confirm classification and manipulation outcomes at a glance as an alternative to the web-based dashboard.

\section{Manipulation Performance and Navigation State-Machine Details}
\label{app:manipulation-performance}

This appendix provides additional manipulation-performance results and implementation details for the navigation state machine. It reports per-trial outcomes for the two proof-of-concept manipulation tasks and describes the finite-state logic coordinating row following, row switching, and recovery behaviors during autonomous navigation.

Table~\ref{tab:submodule-performance} reports the per-trial manipulation outcomes. T/A/S denotes targets available, actions attempted, and actions completed successfully, respectively; Wrong/misplaced records incorrect interactions or misplaced releases, and Time gives the mean execution time per attempted action. For plant toppling, 43 of 45 available targets were attempted, with one left unattempted in Trials 2 and 5. All attempted topplings succeeded with no incorrect interactions, yielding a 100\% attempted-action success rate.

All 55 seed-dispensing targets were attempted, of which 52 succeeded (94.5\%). The three unsuccessful actions comprised misplaced releases in Trials 1 and 3 and an incomplete dispense in Trial 4 caused by a mechanical fault that left the target without a seed.

\begin{table}[H]
\centering
\footnotesize
\caption{Per-trial performance of the two proof-of-concept manipulation tasks over five trials.}
\label{tab:submodule-performance}

\resizebox{0.95\columnwidth}{!}{%
\begin{tabular}{lccccc}
\toprule
\textbf{Task} & \textbf{Trial} & \textbf{T/A/S} &
\makecell{\textbf{Success}\\\textbf{(\%)}} &
\makecell{\textbf{Wrong/}\\\textbf{misplaced}} &
\makecell{\textbf{Time}\\\textbf{(s)}} \\
\midrule

\multirow{6}{*}{\makecell[l]{Plant\\toppling}}
& 1 & 9/9/9 & 100.0 & 0 & 3.61 \\
& 2 & 9/8/8 & 100.0 & 0 & 3.17 \\
& 3 & 9/9/9 & 100.0 & 0 & 3.14 \\
& 4 & 9/9/9 & 100.0 & 0 & 3.30 \\
& 5 & 9/8/8 & 100.0 & 0 & 3.45 \\
& \textbf{Mean} & -- & \textbf{100.0} & -- & \textbf{3.33} \\
\midrule

\multirow{6}{*}{\makecell[l]{Seed\\dispensing}}
& 1 & 11/11/10 & 90.9 & 1 & 10.77 \\
& 2 & 11/11/11 & 100.0 & 0 & 9.84 \\
& 3 & 11/11/10 & 90.9 & 1 & 8.25 \\
& 4 & 11/11/10 & 90.9 & 0 & 10.20 \\
& 5 & 11/11/11 & 100.0 & 0 & 9.80 \\
& \textbf{Mean} & -- & \textbf{94.5} & -- & \textbf{9.77} \\
\bottomrule
\end{tabular}%
}
\end{table}

The manipulation actions were executed within the complete autonomous navigation sequence rather than as isolated subsystem tests. A hierarchical finite-state machine coordinated row following, adjacent-row switching, and recovery behaviors above the continuous controller described in Section~\ref{sec:autonomous-navigation}. RailNav provided the navigation centerline during normal traversal, while range sensing independently detected row ends and initiated row switching. Safety-Align, Camera-Search, and Obstruction-Recovery states handled failed row acquisition, sustained centerline loss, and the asymmetric course break, respectively.

This state-based coordination allowed the robot to repeatedly reach the manipulation locations, execute the required interactions, and continue through the testbed. As reported in Section~\ref{sec:integrated-performance}, 98\% of the integrated trials reached the end of the course; the only incomplete run traversed approximately 85\% of the track.

\end{document}